# A cable-driven parallel manipulator with force sensing capabilities for high-accuracy tissue endomicroscopy

Kiyoteru Miyashita · Timo Oude Vrielink · George Mylonas

**Abstract** This paper introduces a new surgical end-effector probe, which allows to accurately apply a contact force on a tissue, while at the same time allowing for high resolution and highly repeatable probe movement. These are achieved by implementing a cable-driven parallel manipulator arrangement, which is deployed at the distal-end of a robotic instrument. The combination of the offered qualities can be advantageous in several ways, with possible applications including: large area endomicroscopy and multi-spectral imaging, micro-surgery, tissue palpation, safe energy-based and conventional tissue resection. To demonstrate the concept and its adaptability, the probe is integrated with a modified da Vinci robot instrument.

**Keywords** force sensing · endomicroscopy · cable-driven parallel mechanisms · autonomous scanning.

## 1. Introduction

The da Vinci (Intuitive Surgical Inc., CA, USA) is a notable surgical robot which is being used to facilitate complex minimally invasive surgery (MIS). Although the da Vinci is admittedly the most successful surgical robotic system to date, it suffers from a number of limitations. For instance, the lack of haptic feedback has not been satisfactorily resolved despite multiple attempts. Additionally, the instruments offer limited positional resolution for certain applications [1].

On the other hand, there is a demand from the surgeons to be able to utilize the imaging modality intraoperatively for accurate identification of malignant tissue to minimize unnecessary damage to healthy tissue. Optical biopsy using endomicroscopy (EM) provides the highest resolution images among current

Kiyoteru Miyashita
HARMS Lab, Department of Surgery & Cancer, Imperial College London, UK
E-mail: {kiyoteru.miyashita12, t.oude-vrielink15, george.mylonas}@imperial.ac.uk

modalities and is potentially applicable to robotic surgeries as shown in [2]. EM relies on sub-millimetre diameter optical probes to obtain histological information. For accurate and reliable assessment of pre-cancerous lesions and their margins, large areas of tissue must be scanned and accurately reconstructed. The use of a large number of spatially distributed points is one way to cover wide tissue areas. At the same time, the force applied on the tissue by the scanning probe must remain within a very tight range for optimal signal acquisition and safety [3]. For the same reasons, the probe must be able to comply with tissue morphological variations. Research is being conducted and outcomes have been reported by measuring the contact force using a force sensor placed at the tip of an instrument [4], by using force adaptive device [5], or by using computer vision techniques such as edge detection or intensity recognition [6]. However, for the purpose of acquiring high quality images and to avoid a damage to a tissue surface, accurate positional control of the probe and the detection of small contact forces is required.

An alternative and simpler approach is proposed here for robotic EM. This is based on the use of a cable-driven parallel manipulator (CDPM). The proposed mechanism is used to accurately control an EM probe and at the same time accurately measure tissue contact forces. The contact forces are derived simply by measuring the tension of the controlling cables/tendons. As it will be shown, and as already known from existing literature, CDPMs exhibit several desirable qualities. These include: high force transmission, large workspace, easy workspace reconfigurability, high dynamic capabilities, and low cost [7]. This paper is dedicated to the description of a novel instrument based on the CDPM principle. To demonstrate its potential and favourable characteristics, the instrument is retro-fitted on a modified da Vinci robot instrument. Partial robotic hepatectomy is used as an exemplar target procedure, where EM can be very useful in identifying the malignant tissue margins intraoperatively [8, 9]. Therefore, the system is validated on a bovine liver tissue. The following sections provide a description of the developed system and its principle of operation and conclude with experimental validation of its capabilities.

## 2. Instrument Description

In this section, the concept of CDPM is introduced first. Then the design and specification of the new instrument is provided.

### 2.1. Cable-Driven Parallel Robotic Manipulator

The proposed instrument is based on a patented concept introduced in [10], named 'CYCLOPS', which describes a robotic attachment for endoscopes (Fig.1). The CYCLOPS is based on a six-cable CDPM used to control an over-tube that can accommodate commercially available surgical instruments. The CDPM is supported within a deployable rigid or inflatable scaffold, which then turns rigid or semi-rigid scaffold upon deployment or inflation.

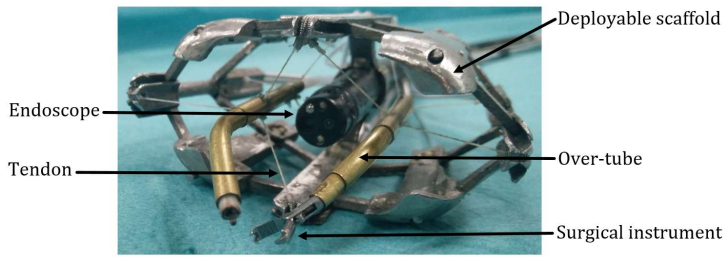

**Fig. 1.** CYCLOPS endoscope robotic attachment

The CYCLOPS exhibits several desirable properties, which are inherent to most CDPM approaches. These include, dexterous manipulation, high accuracy, stability, simplicity, an economic design, and end-effector force sensitivity which is a function to detect a slight force [10]. Existing implementations are based on 3D printed components, polymer tendons, and off-the-shelf accommodated surgical instruments. The system is currently undergoing pre-clinical *in vivo* validation with a focus on colorectal Endoscopic Submucosal Dissection (ESD). Here, we use EM as an exemplar application to further highlight the advantages of the concept.

**2.2. Conceptual design of the proposed instrument**

Unlike the six tendons per instrument used in the CYCLOPS, the proposed instrument uses four tendons to actuate a cylindrical over-tube with an outer diameter of 3.0mm. The over-tube is used to accommodate an EM probe with an outer diameter of 2.8mm. The four tendons can actuate the probe in three planar degrees of freedom (DoF). Fig.2 shows a schematic illustration of the concept.

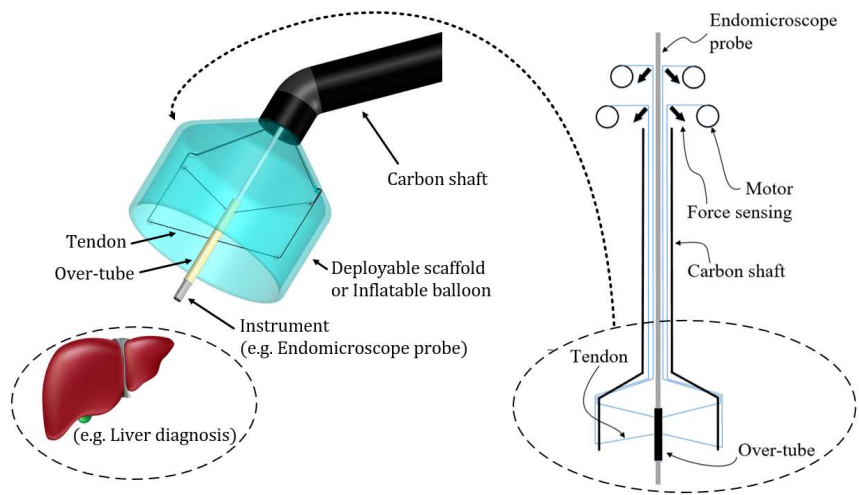

**Fig. 2.** Conceptual design of the proposed instrument

## 2.3. Force Calculation and Sensitivity

With reference to Fig.2, the four-tendon configuration can be used to control the position of the probe and simultaneously serve the purpose of sensing the forces applied on it. The forces on the tip of the probe are estimated by measuring the tension $T_i$ for each tendon $i$, as shown in Fig.3a and (1). The calculated contact force $CF_{cal}$ is derived from the instrument pose and the measured tensions, as shown in Fig.3b and (2). The force sensitivity can be estimated by comparing the calculated forces $CF_{cal}$ with 'ground truth' forces CF, measured with a loadcell placed at the interface with the probe tip. Discrepancies between $CF_{cal}$ and $CF$ are expected due to static friction, elasticity, and hysteresis in the mechanism. $T_0$ is an initial preset tension to keep the over-tube stable before applying the external force.

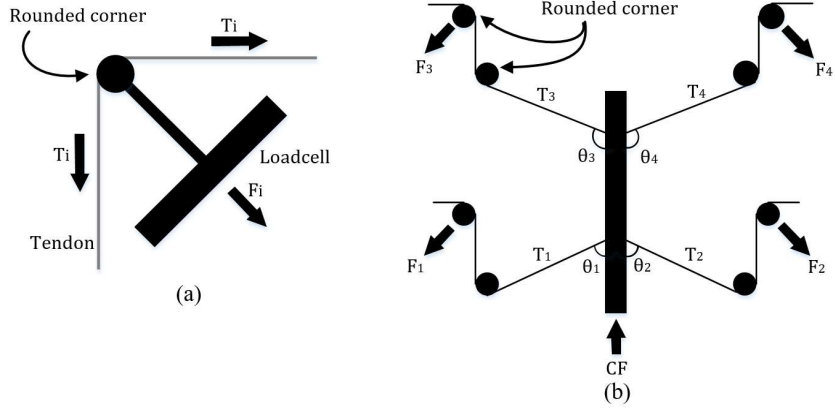

**Fig. 3.** (a) Tension measurement per tendon and (b) Contact force calculation.

$$T_i = \frac{F_i}{2 \sin(\frac{\pi}{4})} \quad (1)$$

$$CF_{cal} = \sum_{i=1}^{n}(T_i - T_{i,0}) \cos \theta_i \quad (2)$$

Depending on the tissue type, the required contact force can be under 1.0N [11]. The reduction of the friction is essential to increase the forces sensitivity of the system, especially for the small contact forces required for proper EM imaging. To reduce friction on the rounded corner where the tendons change directions on the scaffold, or where they interface with the load cells, we use Polytetrafluoroethylene (PTFE) tubes. Using pulleys is an alternative, which however takes more space and introduce more complexity. Moreover, through experimentation we have established that the difference between pulleys and PTFE tubes in terms of friction is negligible. For the proposed device, the tendons in use are 0.19mm diameter UHMW-PE spectra wires (PowerPro, Shimano, Inc., Japan), which can withstand up to 13kgf.

## 2.4. Mechanical design of proposed instrument

All parts are mounted on a modified da Vinci instrument. Fig.4 shows the instrument. Four DC servo motors, 2232S024BX4 CCD 3830 (FAULHABER inc., Daimlerstrabe, Germany) are used to control the tendons. With 3000 increments per revolution, a 1:25 gear ratio and a 10mm tendon spool, the theoretical position accuracy for each tendon is 0.4 µm. Load cells (LCL-005, OMEGA Engineering inc, Manchester, UK) are used, which can measure up to 2.27kgf with 2mV/V ± 20% rated output by a 5VDC excitation voltage. The system is controlled by a laptop PC in C++. Most structures have been manufactured in Poly-Lactic Acid (PLA) by a commercial 3D printer (Ultimaker2 Extended, Ultimaker Inc., Netherlands).

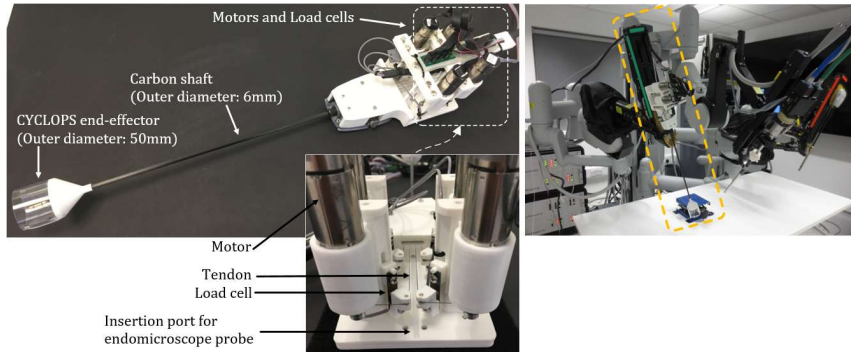

**Fig. 4.** The developed instrument retro-fitted on a standard da Vinci instrument base. On the right, the instrument is shown mounted on da Vinci's slave arm.

## 3. Experimental Validation

In this section, the accuracy of force sensing is discussed by comparing the calculated contact force with the measured contact force in a static situation. Also, the relation between contact forces and acquired EM images is evaluated in a simple 1DoF axial movement. Subsequently, concurrent force sensing and image acquisition in a 2DoF controlled movement is presented.

## 3.1. Measurement of contact force

For the experimental validation of the proposed system, a testing rig has been created. The mechanism to test the contact force is shown in Fig.5. A linear actuator, L2818S0604-T5X5 (Nanotec Electronic inc., Feldkirchen, Germany) is placed at the bottom of the testing mechanism to simulate up-down contacting movement. A load cell (LCL-005, OMEGA Engineering inc, Manchester, UK) is placed on the tip of the linear actuator and used to provide the ground truth. The position of the over-tube is set at the centre of the scaffold, where the tensions in the cables are equal to maintain the static equilibrium condition. The linear actuator is programmed to perform 10 consecutive 2mm displacements, along the longitudinal axis of the probe. Both ground truth and calculated contact forces are acquired and evaluated.

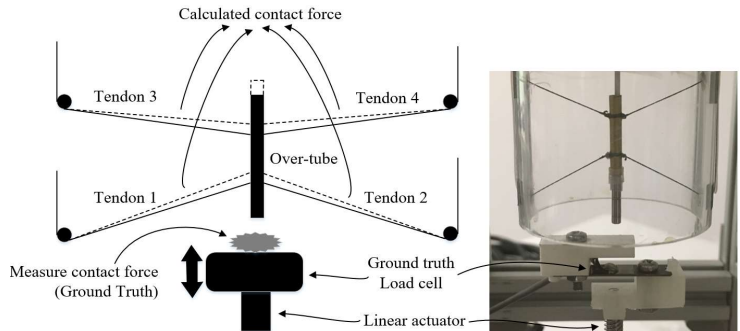

**Fig. 5.** The test-setup to evaluate the contact force sensitivity of the system using a linear stage and ground truth load cell.

## 3.2 Relation of contact force and EM image quality

To assess the capabilities of the system when applied to EM, the optimal applied force for adequate image quality should be assessed and compared to the force sensitivity of the proposed mechanism.

### 3.2.1 Experimental environment

The Cellvizio endomicroscope (Mauna Kea Technologies Inc., Paris, France) is used. An optical probe is inserted from the back-end of the instrument, and placed into the over-tube to which the probe is fixed (Fig.6a). A *Coloflex* probe with a confocal length of 55-65μm is used. The field-of-view diameter of the probe is approximately 0.5mm.

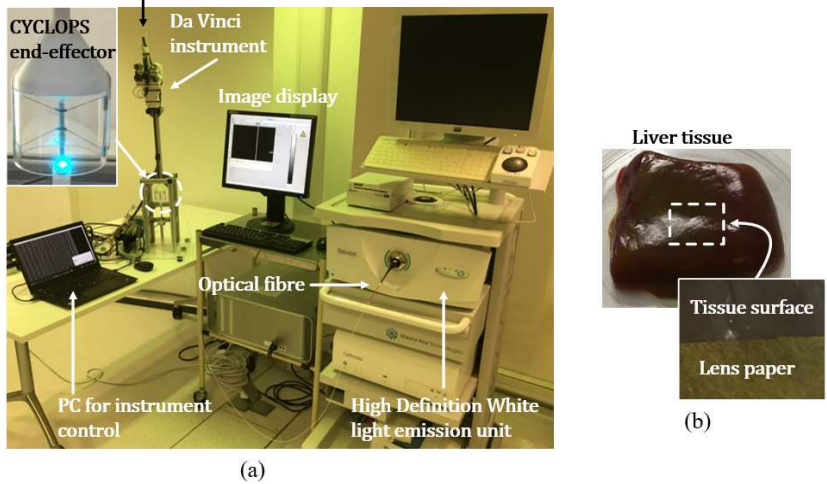

**Fig. 6.** (a) Setup with the endomicroscope system and (b) Liver tissue sample

*3.2.2 Target tissue*

A piece of healthy bovine liver was used as a sample tissue. As the normal liver does not have noticeable features on the surface, a piece of lens-cleaning paper (THORLABS inc., NJ, USA) was placed on top of the liver. The dimension of the tissue is approximately 20×20×8mm, and both the tissue and the paper were dyed with 0.2% acriflavine hydrochloride solution to get a fluorescent effect (Fig.6b).

*3.2.3 Optimum contact force*

As in EM there is a direct relation between the applied force and the image quality achieved, the optimum contact force for the tissue needs to be identified. This contact force can be determined by the ground truth load cell which is placed underneath the liver tissue (Fig.7).

*3.2.4 Control method for image quality assessment*

The over-tube is controlled autonomously by 0.01mm incremental steps in the vertical direction. When the instrument detects contact with an object, the motors are stopped in holding mode. Contact is sensed through a rapid increase of the first derivative on the sensed cable tension values. After this point, the over-tube is moved backwards until the contact force becomes zero again. During this autonomous movement, ground truth force data and EM images are acquired and used for evaluation (Fig.7). Here, to detect the moment that the instrument comes into contact with an object, rigid material (PLA) is initially used instead of the liver tissue. A testing paper was placed on the rigid surface and tested. After experimenting on the rigid surface, the same experiment is repeated with liver tissue.

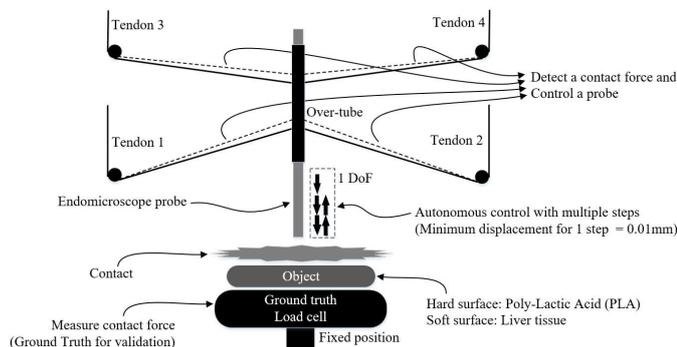

**Fig. 7.** Test-setup used to determine the optimum contact force and determine the maximum force sensitivity in a 1DoF motion.

### 3.3. Autonomous 2DoF control

Autonomous control is essential in a clinical case, because manual control would be difficult to insure stable sub-mm accuracy which is required both for the vertical control based on the confocal length (55-65μm) and for the horizontal control based on the field of view (approximately 0.5mm) of the EM probe. The liver tissue is used for testing the performance in 2 DoF movement using the inverse kinematics of the CDPM, based on a pre-defined surface scanning pathway. All pathways have been preliminarily programmed in C++. The testing mechanism is the same as in section 3.2. The difference is that we use the second derivative of tension data to detect contact force. 2DoF dynamic effects introduce changes in the first derivative even without any surface contact, and therefore additional threshold based on the second derivative is required to increase robustness of force-detection. Threshold values $P$, $Q_n$, and $R_n$(3, 4, 5) are used. $P$ is a difference of the first derivative comparing the sum of tensions of tendon 1 and 2 with the sum of tensions of tendon 3 and 4. $Q_n$ is the increase ratio of the first derivative of each tendon. $R_n$ is the difference of the second derivative compared to a reference. $n$ can be 1 to 4 which are the tendon number, and $t_0$ and $t_1$ are reference times to be compared to the current value.

$$P = \sum_{n=1}^{2} \frac{dT_n}{dt} - \sum_{n=3}^{4} \frac{dT_n}{dt} \tag{3}$$

$$Q_n = \frac{dT_{nt}}{dt} / \frac{dT_{nt_0}}{dt} \tag{4}$$

$$R_n = \frac{dT_{nt}^2}{d^2 t} - \frac{dT_{nt_1}^2}{d^2 t} \tag{5}$$

If the detected contact force is higher than the contact force required for obtaining an optimum quality image due to the diminished force-sensitivity caused by the static friction, controlled step-back of the probe can be used to reach the required optimum force. The amount of stepping-back needs to be determined and pre-defined. During the 2DoF autonomous movement, the ground truth force data and EM images are acquired for evaluation (Fig.8). Six horizontal points are used to conduct effectively and repeatedly the same experiments. Consecutive mosaic images can be acquired between each horizontal point. At each point, the same control method as described in previous section is applied, as shown in Fig.8.

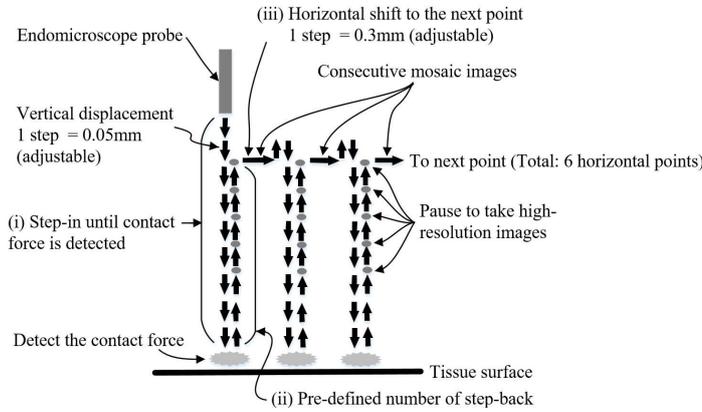

**Fig. 8.** Pathway of endomicroscope probe when the contact force sensitivity is lower than the optimum contact force for adequate images.

Kiyoteru Miyashita et al.                                                                                9

## 4. Results

### 4.1. Measurement of contact force

Fig.9 shows the plots acquired by the procedure of section 3.1. The top graph illustrates how the four tensions change depending on the position of the end-effector. The bottom graph is the comparison between the ground truth contact force and the calculated contact force. The dotted area A shows that the calculated contact force is 14.7±2.0% lower than the ground truth contact force. The dotted areas B1 and B2 show the hysteresis effects, in which the tensions did not return to the initial tension even after removing the external force applied by the actuator. It is considered that this discrepancy is due to static friction between the tendons and the PTFE and by the elasticity of the structure. However, the overall performance is significantly stable over repeated motions, which demonstrates the high repeatability of the mechanism.

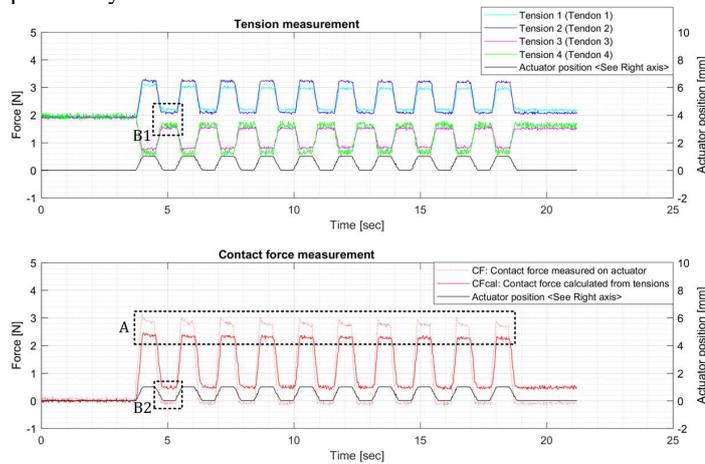

**Fig. 9.** Contact force comparison. Area A shows the error between the calculated and ground-truth contact force. Areas B1 and B2 indicate a discrepancy between measured and calculated contact forces, as a result of hysteresis.

### 4.2. Relation of contact force and EM image

Fig.10 shows EM images obtained using different contact forces. The contact force was measured by a load cell placed under the liver tissue. The image at 0.05N shows the most noticeable contrast with clear edge features. This value was considered as the optimum contact force for acquiring high quality images during EM of the liver tissue in use.

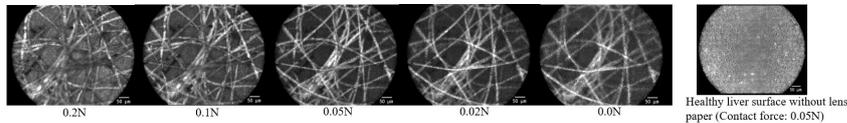

**Fig. 10.** Contact force and respective endomicroscope image

Fig.11 illustrates detection of the contact force and a relation between the contact force and images using the testing paper on a rigid surface. The moving average method was used to reduce the noise of the four tension values. Subsequently, by using the least square method, gradient values were derived as the first derivative. The red line shows the ground truth contact force which was measured on the rigid surface where the testing lens paper was placed. It is noticeable that the contact force was detected at approximately 0.2N, at which point motor actuation was stopped. 0.2N is the minimum contact force which the instrument could detect. This minimum force depends mainly on threshold conditions required to robustly detect forces. At the 0.2N detecting point, the EM image shows certain features of the testing paper, but the image is blurred (Fig.11(i)), as this value is higher than the earlier determined 0.05N force required for acquiring the highest quality EM images. From this point, the EM probe was gradually stepped back from the paper at 0.01mm steps, until the optimum force of 0.05N was reached and the highest quality image was acquired (Fig.11(ii)). The number of back-steps required to reach this optimum force was recorded. The same 0.2N force sensitivity has been found while performing the 1DoF control on the liver tissue.

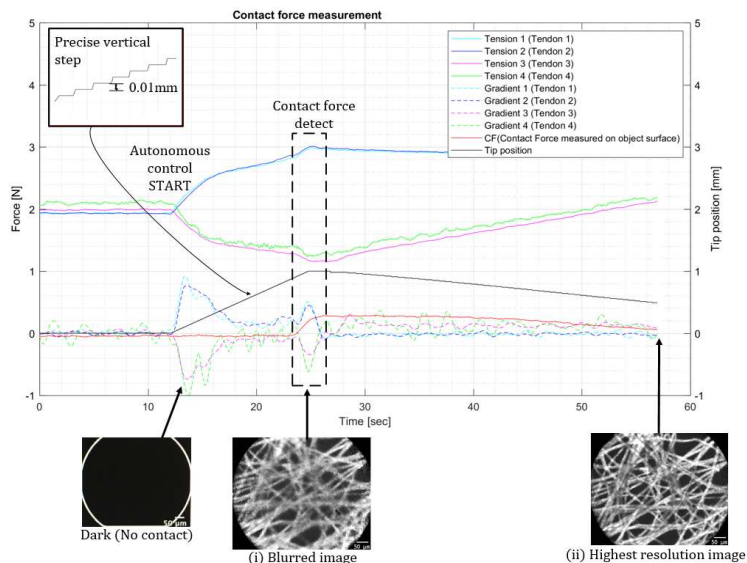

**Fig. 11.** Contact force detection and endomicroscope image

### 4.3. Autonomous 2DoF control

For the 2DoF case, the average minimum detectable contact force on healthy liver tissue was approximately 0.6N, which was larger than the 0.2N found in the 1DoF case due to the difference of the detecting algorithm. Similar to the 1DoF controlled case, the minimum detectable contact force is higher than the required optimum contact force. Therefore, continuous images were taken while stepping back. The pathway in 2DoF is shown in Fig.8, and Fig.12 illustrates the first three points of the 2DoF movement extracting tension data and the first derivative data. *CF* (ground truth contact force measured on the object surface) in Fig.12 shows that approximately 0.6N contact force was detected during each stepping-in motion. Below the graph, representative images are shown. At the detecting point, EM image shows certain features of the testing paper, but the image was blurred (Fig.12(i)). When the contacting force was approximately 0.05N, the highest quality images were taken (Fig.12(ii)). Additionally, clear short mosaic images were also taken during the horizontal movement to move the probe to the next point (Fig.12(iii)). The 2DoF point scanning continued for six points in total. Additional three cases showed minor differences in average detected contact force (0.62N) when compared to the first case, and the optimum contact force (0.04N) when compared to the determined force (0.05N). However, there were no visible negative effects on the EM image quality.

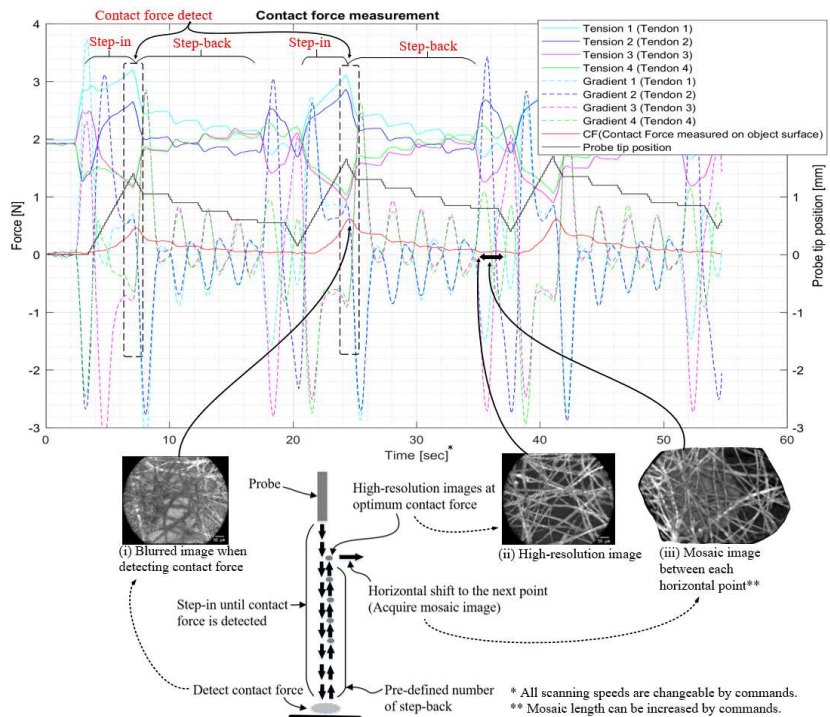

**Fig. 12.** 2DoF autonomous scanning control

## 5. Discussion

This paper shows a first proof-of-concept using a CDPM for safe and adequate contact force to acquire high-resolution EM images. While the detectable force sensitivity is high, the used healthy liver sample has shown to require an even higher force sensitivity for high-quality images. However, the proposed back-stepping technique effectively shows that the positional accuracy of the system is sufficiently high to acquire high-quality images. Additional bench-marking is required to fully validate the positional feed-forward accuracy of the end-effector. Hardware and software improvements can be implemented to further decrease friction and increase the detection sensitivity. It is noteworthy to be mentioned that current results are found without any additional modeling in which the theoretical optimum tension distribution, dynamic and friction effects are taken into account. Implementing such models would further increase the force sensitivity of the system.

The large variability between different tissue pathologies, different tissue types, and different patients will make an adequate solution that is based on contact force alone as a metric for image quality unlikely. The combination of the current method and existing computer vision techniques would enable the system to autonomously and safely obtain high-quality EM images over large tissue sections. In such a case, no information about the tissue is required a priori to attain the optimum contact force for adequate images. Further efforts should be conducted to incorporate the current system with computer vision techniques, and explore the most effective algorithm for large area scanning.

Further hardware developments will include a system with additional cables to allow for 5Dof scanning of larger and complex surface area. Additional integration with the da Vinci and hybrid control methods can be used to increase the workspace of the instrument further. For clinical applicability, a deployable scaffold or an inflatable balloon needs to be implemented. When inserting the instrument into the patient body, an outer diameter of folded scaffold or deflated balloon should be minimised, which would be the same as or near the diameter of the carbon shaft.

While the research focuses primarily on EM imaging, it should be noted that the high force sensitivity can also be used for other clinical applications, including accurate control of a laser scalpel or a micro ultrasound probe. Haptic feedback and palpation are two other features that the current system can be used for, giving the surgeon enhanced situational awareness and diagnostic tools.

## 6. Conclusion

An integration of a CDPM and a da Vinci instrument was introduced providing a viable solution for the safe and precise autonomous EM scanning. 0.2N and 0.6N force sensitivity were found for 1DoF and 2DoF image acquisition methods, respectively. In case of insufficient force sensitivity for specific tissues, the system is accurate enough to acquire high-quality images by gradually stepping back. As force sensing and accurate control have been technical challenges since the emergence of the surgical robots, this new instrument could offer an option not only for robotic EM but also for other applications which require force sensing and precise control.

**Acknowledgments**

We would like to thank Alexandros Kogkas, Ming Zhao, Fernando Avila-Rencoret and Dr Stamatia Giannarou for their help with this research, and Lin Zhang and Khushi Vyas for their support with the da Vinci and Cellvizio systems respectively.